\documentclass{article}
\usepackage{ijcai25}

\usepackage{times}
\usepackage{soul}
\usepackage{url}
\usepackage[hidelinks]{hyperref}
\usepackage[utf8]{inputenc}
\usepackage[small]{caption}
\usepackage{graphicx}
\usepackage{amsmath}
\usepackage{amsthm}
\usepackage{booktabs}
\usepackage{algorithm}
\usepackage{algorithmic}

\usepackage{hyperref}
\usepackage{url}

\usepackage{amssymb}
\usepackage{xcolor}
\usepackage{comment}
\usepackage{enumitem}
\usepackage{threeparttable}
\usepackage{amsmath}

\usepackage{multirow}

\usepackage{graphicx}
\usepackage{subcaption}
\usepackage[multiple]{footmisc}

\usepackage{caption}
\title{UniST-Pred: A Robust Unified Framework for Spatio-Temporal Traffic Forecasting in Transportation Networks Under Disruptions}

\author{
Yue Wang$^1$
\and
Djellel Difallah$^1$\and
Areg Karapetyan$^{1}$\And
Samer Madanat$^1$\\
\affiliations
$^1$New York University Abu Dhabi\\
\emails
\{yw3576, dd115, ak7550, smm17\}@nyu.edu
}
\begin{document}

\maketitle

\begin{abstract}

%and supports

%In real-world deployments, forecasting models must operate under structural uncertainty, sensor sparsity or failure, and partial network disconnection, conditions that are rarely considered in model design.

%UniST-Pred leverages feature-time mixing for efficient temporal dependency learning, learns task-adaptive graph structures for spatial encoding, and fuses both via a lightweight squeeze-excitation residual block. 

Spatio-temporal traffic forecasting is a core component of intelligent transportation systems, supporting various downstream tasks such as signal control and network-level traffic management. In real-world deployments, forecasting models must operate under structural and observational uncertainties, conditions that are rarely considered in model design. Recent approaches achieve strong short-term predictive performance by tightly coupling spatial and temporal modeling, often at the cost of increased complexity and limited modularity. In contrast, efficient time-series models capture long-range temporal dependencies without relying on explicit network structure. We propose UniST-Pred, a unified spatio-temporal forecasting framework that first decouples temporal modeling from spatial representation learning, then integrates both through adaptive representation-level fusion. To assess robustness of the proposed approach, we construct a dataset based on an agent-based, microscopic traffic simulator (MATSim) and evaluate UniST-Pred under severe network disconnection scenarios. Additionally, we benchmark UniST-Pred on standard traffic prediction datasets, demonstrating its competitive performance against existing well-established models despite a lightweight design. The results illustrate that UniST-Pred maintains strong predictive performance across both real-world and simulated datasets, while also yielding interpretable spatio-temporal representations under infrastructure disruptions. The source code and generated dataset are publicly available at {\url{https://github.com/yw3576/UniST-Pred}}.

%Across both real-world and simulated datasets, the results demonstrate that UniST-Pred maintains strong predictive performance and yields interpretable spatio-temporal representations under infrastructure disruptions. 

%Experiments across multiple traffic benchmarks demonstrate competitive accuracy with a lightweight and modular design. We further evaluate UniST-Pred in severe network disconnection scenarios using a realistic traffic simulator, demonstrating robust performance and interpretable spatio-temporal representations after infrastructure disruption.

\end{abstract}

\section{Introduction}
\label{sec:intro}

%Why is this work important for social good? Mandated Information To Be Included in Submissions: All submissions should clearly state in the main text explicitly what real-world societal problem is being tackled, who (including the stakeholders/partners and multi-disciplinary domain experts) is and how they are participating (e.g., their roles) in the project/paper, what the proposed methodologies are, how the methodologies are evaluated (e.g., metrics), and what the real-world impact of the methodologies is. For example, active collaboration with stakeholders/partners and multi-disciplinary domain experts can be explicitly documented in paragraphs near the contributions or in separate sections/subsections and emphasized throughout the main submission when appropriate.

Urban traffic systems operate as large-scale, tightly coupled dynamical networks, where local disturbances can quickly propagate across space and time to produce corridor- or network-level congestion. In many traffic control paradigms \cite{wei2019rlsurvey}, an implicit but critical requirement is \emph{anticipation}: decisions are made based not only on current observations, but also on expected near-future traffic evolution. Traffic forecasting is central to operational traffic management, enabling agencies to respond proactively to localized infrastructure impediments such as bridge closures, flooded road segments, or zoned capacity reductions \cite{li2022nonlinear,manibardo2021deep}. By accounting for such disruptions, traffic controllers can prevent cascading congestion effects and maintain stable network operation under non-recurrent events that significantly impact mobility and safety. However, most existing forecasting models assume complete and stable network structure, limiting their practical applicability in real-world deployments. \looseness-2

In this work, we address this gap by rethinking the design of the traffic forecasting module around properties that directly support effective traffic control, including (i) low-latency and lightweight inference suitable for deployment in traffic control centers, (ii) modular integration with existing controllers, and (iii) interpretable spatio-temporal representations that disentangle spatial and temporal features, enabling practitioners to reason about evolving traffic conditions.

Research on traffic forecasting has increasingly focused on deep spatio-temporal models. Existing state-of-the-art approaches can be broadly grouped into two representative directions.
The first line focuses on \emph{spatio-temporal graph neural networks and transformers} \cite{jiang2023pdformer,wang2024stgformer}, which couple temporal modeling and spatial interaction learning through stacked message passing, recurrent units, or attention mechanisms. Representative methods achieve strong predictive accuracy by modeling long-range and dynamic dependencies on traffic networks. However, these models often rely on intertwined spatio-temporal blocks, leading to increased architectural complexity and computational cost. Such tight coupling also reduces modularity, making it less convenient to adapt or integrate the predictor as a standalone component in larger systems.

The second line emphasizes \emph{efficient temporal modeling} in time-series forecasting \cite{chen2023tsmixer,nie2022time,liu2024itransformer}. Recent mixer-style or transformer-based backbones demonstrate that long-range temporal dependencies can be captured effectively without heavy recurrent or attention-based sequence modeling. While these approaches offer favorable scalability, they typically treat each location independently or rely on simple spatial features, and thus do not directly address how to incorporate network structure in a task-adaptive manner. \looseness-2

Most existing methods emphasize either developing expressive spatio-temporal coupling or efficient temporal modeling, but few attempt to address both within a unified and modular design. This motivates a structural reconsideration of how temporal and spatial information should be modeled and combined for traffic forecasting.
We propose \textbf{UniST-Pred}, a unified spatio-temporal forecasting framework built around a simple but deliberate design principle: \emph{decouple temporal dependency learning from spatial representation extraction, and then fuse them adaptively at the representation level}. By separating temporal modeling, spatial encoding, and fusion into distinct stages, UniST-Pred departs from tightly coupled spatio-temporal architectures, offering a unified yet compositional alternative for traffic forecasting. %Rather than introducing new control mechanisms or deeply interleaving spatio-temporal propagation,
%UniST-Pred aims to provide  \looseness-2

Leveraging the proposed framework, we assemble a competitive, lightweight traffic forecasting model suitable for integration into traffic analysis and control pipelines. The temporal component of the model employs the feature-time mixing technique proposed in \cite{chen2023tsmixer} to capture long-range temporal dependencies and cross-feature interactions without recurrent or attention-based sequence modeling. The spatial component follows the idea of Graph Transformer Networks (GTNs)~\cite{yun2019graph}, learning task-adaptive graph structures by composing multiple candidate relations before applying graph convolution. Finally, temporal and spatial representations are integrated through a squeeze-and-excitation \cite{hu2018squeeze} residual fusion block, which adaptively reweights the two feature streams while preserving modularity. Overall, the contributions of this work are four-fold:
\begin{itemize}[leftmargin=*, label=$\star$]
    \item \textbf{A unified decouple-then-fuse spatio-temporal forecasting framework:}
    We present a modular modeling paradigm that separates temporal dependency learning from spatial representation extraction and integrates them through adaptive representation-level fusion, providing a competitive and adaptive alternative to tightly intertwined spatio-temporal architectures.

\item \textbf{A curated simulator-based traffic forecasting dataset with structural network variations:} We provide a traffic flow dataset, referred to as \textbf{SimSF-Bay}, derived from MATSim~\cite{w2016multi} simulations of the San Francisco (SF) Bay area's transportation network (see Sec.~\ref{sec:matsim_dataset} for a brief overview of MATSim).  The dataset comprises various disrupted-network scenarios generated by removing selected roads/bridges from the underlying MATSim network, with each scenario corresponding to a specific combination of these roads and/or bridges. These scenarios induce changes in network connectivity and topology, enabling evaluation of model robustness under structural/observational uncertainties that are not captured by existing standard benchmarks.

%The dataset incorporates scenario-dependent network disruptions and topological changes. SimSF-Bay enables evaluation of model robustness under structural/observational uncertainties that are not captured by existing standard benchmarks. \looseness-2

    \item \textbf{A lightweight and robust traffic forecasting model:} 
We develop a parameter-efficient traffic forecasting model within the proposed framework and demonstrate its robustness under a range of network disruption scenarios reflecting real-world events such as road closures, incidents, or infrastructure contingencies.
    \item \textbf{Thorough empirical evaluation:} We conduct an extensive evaluation of the presented model, including quantitative and qualitative performance comparison against well-established traffic forecasting methods on two widely used benchmark datasets, PEMS-Bay \cite{li2017diffusion} and NYCTaxi \cite{zheng2014nyctaxi}. Lastly, we examine model interpretability and provide insights into the learned spatio-temporal representations under simulated network disruptions. \looseness-2   
\end{itemize}

\paragraph{Real-World Value and Collaboration.}
This work is inherently multidisciplinary, bringing together experts from transportation systems, infrastructure planning, and machine learning communities, including the last author, who is a senior transportation engineer and scientist with first-hand knowledge of the studied San Francisco Bay network. With over 20 years of experience, the last author has contributed to multiple projects and collaborations with transportation agencies both within and outside the U.S., including the Bay Area Metropolitan Transportation Commission (MTC), and is currently leading an active project with the federal infrastructure agency of the UAE. In this work, they helped shape the central motivation of studying traffic forecasting under network disruptions and provided domain expertise for the design and qualitative assessment of the disruption case studies.

\section{Related Work}
\label{sec:related}

\subsection{Spatio-Temporal Traffic Forecasting with Coupled Architectures}
Most existing deep learning approaches for traffic forecasting adopt \emph{coupled} spatio-temporal modeling strategies \cite{jiang2023gnn_survey,yu2018stgcn,li2021stfgnn}, where spatial interaction learning and temporal dependency modeling are tightly intertwined. 
Early models such as STGCN \cite{yu2018stgcn} combine graph convolution with temporal convolution to capture localized spatial-temporal dependencies, while DCRNN \cite{li2017diffusion} models traffic diffusion on directed graphs using recurrent encoder-decoder structures. Graph WaveNet \cite{wu2019graph} further improves this line by introducing an adaptive dependency matrix to capture hidden spatial relations beyond predefined road connectivity. \looseness-2

More recent transformer-style architectures further strengthen this coupling by using spatio-temporal attention to dynamically model interactions across nodes and time steps \cite{zheng2019gman,wang2024stgformer,jiang2023pdformer}. ASTGCN \cite{guo2019attention} leverages spatial and temporal attention to identify important nodes and time intervals. STFGNN \cite{li2021stfgnn} builds a spatial-temporal fusion graph to capture correlations across nodes and time steps. AGCRN \cite{bai2020adaptive} and MTGNN \cite{wu2020connecting} learn adaptive graph structures for end-to-end spatial-temporal forecasting. Transformer-based models, such as GMAN \cite{zheng2019gman}, STAEformer \cite{liu2023spatio}, PDFormer \cite{jiang2023pdformer}, and STGformer \cite{wang2024stgformer}, further improve long-range dependency modeling through spatial-temporal attention, adaptive embeddings, propagation-aware mechanisms, and efficient graph transformer designs. These models demonstrate strong predictive accuracy on large-scale benchmarks, particularly when long-range spatial and temporal dependencies are prominent. However, their performance often comes at the cost of increased architectural complexity and computational overhead. More importantly, the repeated interleaving of spatial interaction learning and temporal modeling reduces modularity: spatial and temporal components are difficult to isolate, reuse, or simplify, which can hinder flexible deployment when forecasting models are used as standalone modules in larger systems. \looseness-2

\begin{figure*}[h]
    \centering
    \includegraphics[width=1\textwidth]{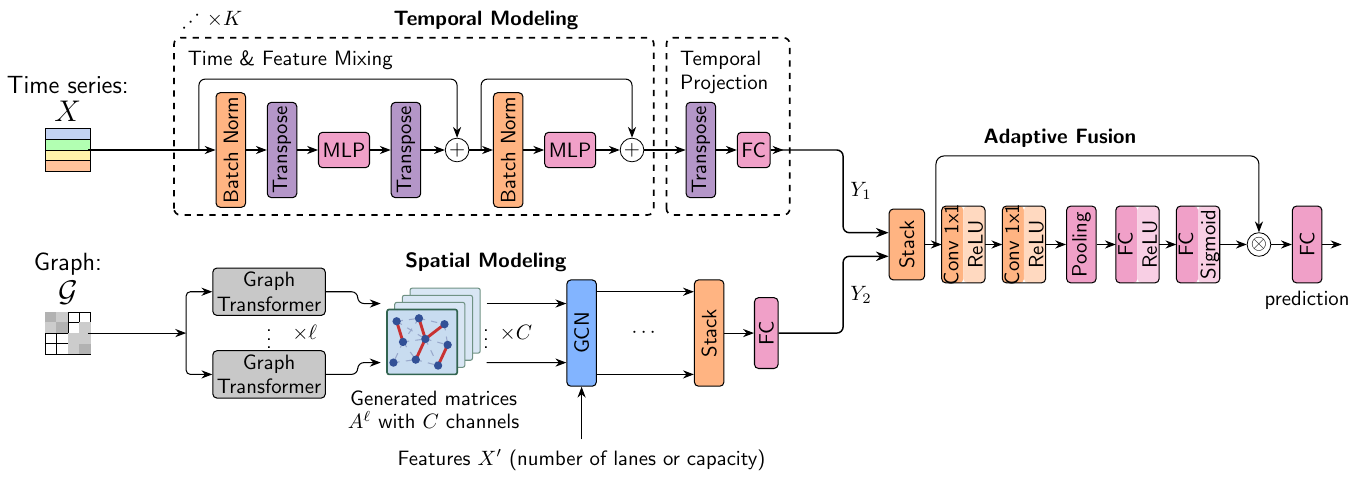} % Do not include the file extension
    \caption{The architecture of the proposed UniST-Pred model.}
    \label{fig:methodology}
\end{figure*}

\subsection{Decoupled Perspectives: Efficient Temporal Modeling and Task-Adaptive Spatial Encoding}
Recent studies have explored more modular or decoupled learning strategies for traffic forecasting. ST-SSL \cite{ji2023spatio} improves spatial-temporal representations through self-supervised objectives, partially separating representation learning from the final supervised forecasting task. Along this direction, STEP \cite{shao2022pre} and STD-MAE \cite{gao2024spatial} further demonstrate the value of pre-training-based representation learning. STEP adopts a pre-training and fine-tuning paradigm to learn transferable spatial-temporal knowledge before adapting it to downstream forecasting tasks. STD-MAE extends this perspective by explicitly decoupling spatial and temporal dependency learning within a masked pre-training framework. Together, these methods illustrate that separating representation acquisition from task-specific prediction can improve transferability and robustness in spatial-temporal forecasting.

In parallel, the time-series forecasting community has shown that long-range temporal dependencies can be modeled efficiently without heavy recurrent or attention-based sequence modeling. Mixer-style architectures \cite{chen2023tsmixer} and efficient transformer variants \cite{nie2022time,zhang2024patchtcn} demonstrate that temporal dependency learning can be decomposed into simple operations along the time and feature dimensions, achieving competitive performance with significantly reduced complexity. These results suggest that expressive temporal modeling does not inherently require tight coupling with spatial interaction mechanisms.

Separately, another line of research questions the reliance on fixed graph structures for spatial modeling. Graph structure learning and relation composition methods argue that a single predefined adjacency matrix may be suboptimal for downstream tasks. Graph Transformer Networks \cite{yun2019graph} exemplify this perspective by learning task-adaptive graph structures through soft composition of candidate relations. In traffic forecasting, related ideas appear in models that infer dynamic or attention-based spatial dependencies rather than relying solely on physical road connectivity.

\section{Preliminaries \& Problem Statement}
\label{sec:preliminary}
We consider the problem of traffic forecasting on a road transportation network represented as a weighted directed graph
$
\mathcal{G} = (\mathcal{V}, \mathcal{E}, \mathbf{A}),
$
\noindent where $\mathcal{V}=\{v_1,\dots,v_N\}$ denotes the set of road segments or intersections, $\mathcal{E}\subseteq\mathcal{V}\times\mathcal{V}$ is the set of directed edges describing allowable traffic flow, and $\mathbf{A}\in\mathbb{R}^{N\times N}$ is the weighted adjacency matrix encoding spatial relationships such as distance or connectivity strength.

Each node $v_i\in\mathcal{V}$ is associated with a traffic time series 
$
\mathbf{x}_i=[x_i^1,x_i^2,\dots,x_i^T],
$
\noindent where $x_i^t$ denotes the observed traffic variable (e.g., speed, flow, or density) at time $t$. The network state at time $t$ is $\mathbf{X}^t=[x_1^t,\dots,x_N^t]^\top\in\mathbb{R}^N$. In addition, each node may have static features $\mathbf{X}'\in\mathbb{R}^{N\times d}$, such as lane count or capacity.

Given a historical observation window of length $H$, the task is to predict future traffic states for a horizon $H'$:
\[
f_\theta:(\{\mathbf{X}^{t-H+1}, \dots, \mathbf{X}^{t}\}, \mathcal{G}, \mathbf{X}')
\rightarrow
(\hat{\mathbf{X}}^{t+1}, \dots, \hat{\mathbf{X}}^{t+H'})
\]
where $f_\theta(\cdot)$ is the predictive model parameterized by $\theta$. The function $f_\theta$ must jointly capture temporal dependencies in the historical sequences $\mathbf{X}^{t-H+1:t}$ and spatial correlations encoded in the graph $\mathcal{G}$, as traffic dynamics are influenced by both local temporal trends and interactions among neighboring road segments. In practice, temporal feature extraction (e.g., time mixing) is integrated with graph-based representations to model spatio-temporal dynamics, allowing the model to learn complex dependency patterns that evolve over time and vary across different regions of the network, enabling robust and adaptive traffic prediction.

\section{Methodology}
\label{sec:methodology}

We propose a unified spatio-temporal framework for traffic forecasting under dynamic scenarios where network topology may change (e.g., road closures). The model decouples temporal modeling and spatial representation learning, then integrates them via an adaptive fusion module (Fig.~\ref{fig:methodology}). For temporal dependencies, we adopt feature/time mixing \cite{chen2023tsmixer} to capture long-range dynamics efficiently, and for spatial dependencies we employ graph-based encoding to leverage road-network structure without assuming a fixed topology. This separation-and-fusion design improves robustness to evolving connectivity and heterogeneous traffic conditions.
\paragraph{Temporal Modeling via Feature and Time Mixing:}
The temporal block models long-range dependencies and cross-feature interactions using feature/time mixing instead of RNNs or attention. This design captures both short-term variations and long-term patterns efficiently. The time and feature mixing layer is stated as below:
\paragraph{(a) Mixer operators.}
We use $\mathcal{M}$ to denote structured mixing operators applied to tensor
representations.
Specifically, $\mathcal{M}_{\mathrm{time}}$ denotes a time-mixing operator that
acts along the temporal dimension independently for each node and feature,
while $\mathcal{M}_{\mathrm{feat}}$ denotes a feature-mixing operator that models
cross-channel interactions at each time step.
Both operators preserve the input shape and are implemented as learnable
linear-nonlinear transformations. \looseness-2
\paragraph{(b) Stacked temporal operator.}
The temporal modeling function is defined as a composition of $K$ stacked
temporal mixer blocks:
$
f_{\mathrm{temp}}
:=
\mathcal{T}^{(K)}_{\theta_K}
\circ
\mathcal{T}^{(K-1)}_{\theta_{K-1}}
\circ
\cdots
\circ
\mathcal{T}^{(1)}_{\theta_1}
$, where each temporal mixer block $\mathcal{T}^{(k)}_{\theta_k}$ is given by
\begin{equation}
\mathcal{T}^{(k)}_{\theta_k}
=
\mathcal{M}^{(k)}_{\mathrm{feat}}
\circ
\mathcal{M}^{(k)}_{\mathrm{time}},
\qquad k = 1,\ldots,K.
\label{eq:temp_block}
\end{equation}
Given an input sequence $\mathbf{X}_{t-H+1:t}$, the temporal representation is
computed as
\begin{equation}
\begin{split}
\mathbf{Y}_1
= &
f_{\mathrm{temp}}(\mathbf{X}_{t-H+1:t})
= \\
&\left(
\bigcirc_{k=1}^{K}
\mathcal{M}^{(k)}_{\mathrm{feat}}
\circ
\mathcal{M}^{(k)}_{\mathrm{time}}
\right)
\left(\mathbf{X}_{t-H+1:t}\right).
\label{eq:temp_output}
\end{split}
\end{equation}
Here, $\circ$ denotes function composition, and $\bigcirc_{k=1}^{K}$ denotes ordered composition from block $1$ to $K$, with each block applying time-mixing followed by feature-mixing.

By avoiding sequential inductive biases, the temporal module provides a flexible representation of traffic dynamics that generalizes across traffic regimes and temporal resolutions. The resulting temporal embeddings focus on intrinsic time-series characteristics, independent of spatial structure.

\paragraph{Spatial Representation Learning with Graph-Based Encoding:}
The spatial block uses the road network as structural context to extract spatial features. Rather than relying on inductive message passing, it learns which road connections are most relevant to the task. Following a GTN-style approach, the model constructs task-adaptive graph structures by softly masking road segments, combining candidate relations, and applying lightweight graph convolutions.

\paragraph{(a) Relation composition (task-adaptive graph generation).}
For channel $i\in\{1,\ldots,C\}$, we generate an adjacency matrix by
softly composing two selected adjacency matrices $Q^{(l)}_{1,i}$ and $Q^{(l)}_{2,i}$:
\begin{equation}
\mathbf{A}^{(l)}_i
= \mathbf{D}_i^{-1}\left(\mathbf{Q}^{(l)}_{1,i}\mathbf{Q}^{(l)}_{2,i}\right),
\label{eq:gtn_comp}
\end{equation}
where $\mathbf{D}_i$ is the degree matrix of $\mathbf{Q}^{(l)}_{1,i}\mathbf{Q}^{(l)}_{2,i}$.

\paragraph{(b) Graph propagation (per-channel GCN).}
Given node features $\mathbf{X}'\in\mathbb{R}^{N\times d}$ (e.g., lane count or capacity),
we add self-loops and compute normalized propagation:
\[
\tilde{\mathbf{A}}^{(l)}_i = \mathbf{A}^{(l)}_i + \mathbf{I},
\qquad
\tilde{\mathbf{D}}_i \ \text{is the degree matrix of}\ \tilde{\mathbf{A}}^{(l)}_i.
\]
The per-channel representation is
\begin{equation}
\mathbf{H}_i
= \sigma\!\left(\tilde{\mathbf{D}}_i^{-1}\tilde{\mathbf{A}}^{(l)}_i \mathbf{X}' \mathbf{W}\right),
\label{eq:gcn_channel}
\end{equation}
where $\sigma(\cdot)$ is an activation function (e.g., ReLU),
and $\mathbf{W}\in\mathbb{R}^{d\times d}$ is a trainable weight shared across channels.
We concatenate the channel features, $\mathbf{Z} = \big\Vert_{i=1}^{C}\mathbf{H}_i$, and apply a fully connected projection to produce the spatial representation:
\begin{equation}
\mathbf{Y}_2
= f_{\mathrm{spat}}(\mathbf{X}')
= f_{\theta'}(\mathbf{Z}).
\label{eq:spat}
\end{equation}

Unlike deep or recursive graph convolutions, the spatial block is intentionally shallow, enabling stable spatial feature extraction without attempting to model exact network connectivity. This lightweight design produces effective spatial embeddings that capture essential structural patterns and remain robust to partial topology changes.

\paragraph{Adaptive Fusion via Squeeze-Excitation Residual Integration:}
To jointly exploit temporal and spatial information, the proposed framework integrates the two representations through a squeeze–excitation \cite{hu2018squeeze} residual fusion block, which adaptively reweights temporal and spatial features based on their relative importance.

Given the temporal representation
$
\mathbf{Y}_1 = f_{\mathrm{temp}}\!\left(\mathbf{X}_{t-H+1:t}\right),
$
and the spatial representation
$
\mathbf{Y}_2 = f_{\mathrm{spat}}(\mathbf{X}'),
$
we first stack the two representations to form a joint feature:
\begin{equation}
\mathbf{Y} = \mathrm{Stack}(\mathbf{Y}_1,\mathbf{Y}_2).
\end{equation}
The fused representation is finally obtained via a squeeze--excitation residual
mapping:
\begin{equation}
\hat{\mathbf{X}}
=
\mathbf{Y}
+
f_{\theta_{\mathrm{se}}}
\!\left(
f_{\theta_{\mathrm{res}}}(\mathbf{Y})
\right).
\label{eq:fusion_consistent}
\end{equation}
Let $\mathbf{Z}=f_{\theta_{\mathrm{res}}}(\mathbf{Y})$.
The squeeze--excitation function is defined as
\begin{equation}
f_{\theta_{\mathrm{se}}}(\mathbf{Z})
=
\mathbf{Z}\odot
\mathrm{sigmoid}
\!\left(
\mathbf{W}_2
\sigma\!\left(
\mathbf{W}_1\,\mathrm{avg}(\mathbf{Z})
\right)
\right),
\end{equation}
where $\odot$ denotes the Hadamard product.
Here, $\mathrm{avg}(\cdot)$ denotes a global pooling operator that aggregates
$\mathbf{Z}$ along the representation dimension to produce a compact descriptor,
$\sigma(\cdot)$ is a nonlinear activation function (e.g., ReLU), and
$\mathrm{sigmoid}(\cdot)$ maps the gating coefficients to $(0,1)$.
The matrices $\mathbf{W}_1$ and $\mathbf{W}_2$ are trainable parameters that
control the squeeze and excitation stages, respectively.
This formulation adaptively reweights different representation channels while
preserving the spatial and temporal structure encoded in $\mathbf{Z}$.

By performing fusion at the representation level rather than tightly coupling spatio-temporal propagation, the framework maintains modularity and interpretability. The residual connection further ensures stable gradient flow and preserves strong temporal representations while incorporating spatial cues, leading to robust and adaptive traffic forecasting across diverse scenarios.\looseness-2
\paragraph{Final prediction:}
Combining temporal modeling, spatial representation learning, and adaptive
fusion, the output prediction is given by
\begin{equation}
\begin{split}
\hat{\mathbf{X}}_{t+1:t+H'}
&=
\mathrm{Stack}\!\left(
f_{\mathrm{temp}}\!\left(\mathbf{X}_{t-H+1:t}\right),
f_{\mathrm{spat}}(\mathbf{X}')
\right)
+ \\
&f_{\theta_{\mathrm{se}}}
\!\left(
f_{\theta_{\mathrm{res}}}
\!\left(
\mathrm{Stack}\!\left(
f_{\mathrm{temp}}\!\left(\mathbf{X}_{t-H+1:t}\right),
f_{\mathrm{spat}}(\mathbf{X}')
\right)
\right)
\right).
\label{eq:predict_consistent}
\end{split}
\end{equation}
The model is trained end-to-end by minimizing the prediction loss:
\begin{equation}
\min_{\theta}
\;
\mathcal{L}
=
\frac{1}{|\mathcal{T}|}
\sum_{t\in\mathcal{T}}
\ell\!\left(
\hat{\mathbf{X}}_{t+1:t+H'},
\mathbf{X}_{t+1:t+H'}
\right),
\label{eq:loss_consistent}
\end{equation}
where $\theta$ collects all learnable parameters in
$f_{\mathrm{temp}}$, $f_{\mathrm{spat}}$, $f_{\theta_{\mathrm{res}}}$ and
$f_{\theta_{\mathrm{se}}}$. \looseness-2
\begin{table*}[t]
\centering
\begin{threeparttable}
\resizebox{1.0\linewidth}{!}{
\begin{tabular}{llllllllll}
\hline
 & \multicolumn{3}{c}{\textbf{SimSF-Bay}}
 & \multicolumn{3}{c}{\textbf{PEMS-Bay}}
 & \multicolumn{3}{c}{\textbf{NYCTaxi}} \\
\cline{2-4} \cline{5-7} \cline{8-10}
\textbf{Method}
& RMSE & MAE & MAPE (\%)
& RMSE & MAE & MAPE (\%)
& RMSE & MAE & MAPE (\%) \\
\hline
ARIMA        & 8.5 {\small(-58\%)} & 4.68 {\small(-49\%)} & 45.8 {\small(-25\%)} & 6.50 {\small(-35\%)} & 3.36 {\small(-43\%)} & 8.34 {\small(-46\%)} & 43.22 {\small(-69\%)} & 15.09 {\small(-69\%)} & 45.3 {\small(-8\%)} \\
LSTM         & 6.4 {\small(-44\%)} & 4.23 {\small(-44\%)} & 53.6 {\small(-36\%)} & 4.96 {\small(-15\%)} & 2.37 {\small(-19\%)} & 5.70 {\small(-21\%)} & 22.30 {\small(-40\%)} & 10.34 {\small(-55\%)} & 122 {\small(-66\%)} \\
TCN          & 5.2 {\small(-31\%)} & 3.40 {\small(-30\%)} & 45.3 {\small(-25\%)} & 5.11 {\small(-18\%)} & 2.76 {\small(-31\%)} & 5.97 {\small(-24\%)} & 15.88 {\small(-16\%)} & 5.45 {\small(-14\%)} & 42.3 {\small(-2\%)} \\
TS-Mixer     & \underline{3.9} {\small(-8\%)} & 2.60 {\small(-9\%)} & 37.1 {\small(-8\%)} & 4.91 {\small(-14\%)} & 2.69 {\small(-29\%)} & 6.06 {\small(-25\%)} & \underline{14.82} {\small(-10\%)} & \underline{5.20} {\small(-10\%)} & 45.0 {\small(-8\%)} \\
TS-Mixer-ext & 4.0 {\small(-10\%)} & 2.58 {\small(-8\%)} & 37.4 {\small(-9\%)} & 4.90 {\small(-14\%)} & 2.69 {\small(-29\%)} & 6.00 {\small(-25\%)} & 14.82 {\small(-10\%)} & 5.20 {\small(-10\%)} & 45.0 {\small(-8\%)} \\
\hline
STGCN        & 6.2 {\small(-42\%)} & 4.11 {\small(-42\%)} & 50.6 {\small(-32\%)} & 5.69 {\small(-26\%)} & 2.49 {\small(-23\%)} & 5.79 {\small(-22\%)} & 26.52 {\small(-50\%)} & 13.34 {\small(-65\%)} & 58.5 {\small(-29\%)} \\
ASTGCN       & 4.5 {\small(-20\%)} & \underline{2.51} {\small(-6\%)} & 49.8 {\small(-31\%)} & 5.42 {\small(-23\%)} & 2.61 {\small(-27\%)} & 6.00 {\small(-25\%)} & 25.30 {\small(-47\%)} & 9.22 {\small(-49\%)} & 47.6 {\small(-13\%)} \\
ST-SSL       & 8.3 {\small(-57\%)} & 4.39 {\small(-46\%)} & 96.4 {\small(-65\%)} & 5.32 {\small(-21\%)} & 2.25 {\small(-15\%)} & 5.13 {\small(-12\%)} & 36.48 {\small(-63\%)} & 10.02 {\small(-53\%)} & 170 {\small(-76\%)}\\
DCRNN        & 4.0 {\small(-10\%)} & 2.75 {\small(-14\%)} & 37.9 {\small(-10\%)} & \underline{4.74} {\small(-11\%)} & 2.07 {\small(-7\%)} & 4.90 {\small(-7\%)} & 15.84 {\small(-15\%)} & 6.43 {\small(-27\%)} & 60.0 {\small(-31\%)} \\
STEP         & 4.0 {\small(-10\%)} & 2.63 {\small(-10\%)} & \underline{36.9} {\small(-7\%)} & \textbf{4.20} {\small(0\%)} & \textbf{1.79} {\small(+7\%)} & \textbf{4.18} {\small(+8\%)} & 22.62 {\small(-41\%)} & 10.41 {\small(-55\%)} & \textbf{32.6} {\small(+28\%)} \\
\hline
\textbf{UniST-Pred}  & \textbf{3.6} & \textbf{2.37} & \textbf{34.2} & \textbf{4.20} & \underline{1.91} & \underline{4.52} & \textbf{13.39} & \textbf{4.70} & \underline{41.6}\\
\hline
\end{tabular}
}
\begin{tablenotes}[flushleft]
\scriptsize
\item[1] The results for PEMS-Bay evaluate the 12th timestep's predictions; the results for NYC-Taxi evaluate the outflow predictions.
\end{tablenotes}
\end{threeparttable}
\caption{Performance comparison on SimSF-Bay, PEMS-Bay, and NYC-Taxi datasets. The best results are \textbf{bold} and the second best results are \underline{underlined}. The values in the brackets denote the relative performance gap w.r.t. UniST-Pred, computed as $100\% \cdot \frac{M_{\text{UniST-Pred}}-M_{\text{method}}}{M_{\text{method}}}$, where $M$ represents the scores (negative values indicate worse performance).}
\label{tab:comparison}
\end{table*}

\section{Experiments}\label{sec:experiments}

\subsection{Experimental Setup \& Settings}

\paragraph{Datasets:} We evaluate our method on three different datasets:
\begin{itemize}
    \item \textbf{SimSF-Bay}: SimSF-Bay, introduced in this work, is a traffic flow dataset derived from MATSim simulations of the SF Bay area's transportation network. The underlying MATSim model and its configurations were adopted from \cite{PAPAKONSTANTINOU2020102515}. The network comprises 7,709 road segments, with traffic flow recorded at 5-minute intervals. The traffic flow from the previous 45 minutes is used as input for predicting the next time step (prediction horizon $H'=1$).
    \item \textbf{PEMS-Bay} \cite{li2017diffusion}: PEMS-Bay is a traffic speed dataset collected from the California Transportation Agencies (CalTrans) Performance Measurement System (PeMS) for the Bay Area. It contains data from 325 sensors over a 6-month period, from January 1, 2017 to May 31, 2017. Traffic information is recorded at 5-minute intervals. Traffic data from the previous 7 days around the predicted time are used to forecast the next 12 time steps (prediction horizon $H'=12$).
    \item \textbf{NYCTaxi} \cite{zheng2014nyctaxi}: NYCTaxi is a taxi trajectory dataset for New York City, spanning from January 1, 2015 to March 1, 2015. The data are recorded at 30-minute intervals. Traffic flow from the previous 17.5 hours is used to forecast the flow for the next time step (prediction horizon $H'=1$).
\end{itemize}

The PEMS-Bay and NYCTaxi datasets follow the standard setting and assume a fixed transportation network or spatial discretization, with traffic states evolving over time. In contrast, SimSF-Bay captures scenario-dependent topological variations, including changes in network connectivity and link availability, while maintaining a fixed simulation horizon consistent with standard agent-based transport modeling practice. This setting enables the assessment of model robustness under structural network changes, which is not supported by existing benchmarks. Additional details on the datasets are provided in Sec.~\ref{sec:matsim_dataset_2}.

\paragraph{Baselines \& Metrics:}

We compare our method against a diverse set of established time series forecasting approaches, covering classical statistical models, temporal deep learning methods, and spatio-temporal graph neural networks. Specifically, the baselines include ARIMA \cite{kumar2015short}, LSTM \cite{hya2014sequence}, TCN \cite{lea2017temporal}, TSMixer and its extended variant TSMixer-ext \cite{chen2023tsmixer}, as well as several spatio-temporal models, including STGCN \cite{yu2018stgcn}, ASTGCN \cite{guo2019attention}, ST-SSL \cite{ji2023spatio}, DCRNN \cite{li2017diffusion}, and STEP \cite{shao2022pre}. For evaluation, we follow standard practice and report RMSE, MAE, and MAPE, formally defined as follows:

\begin{equation}
\begin{split}
&\mathrm{RMSE}=\sqrt{\frac{1}{N}\sum_{t=1}^{N}\left(y_t-\hat{y}_t\right)^2}, \quad \mathrm{MAE}=\frac{1}{N}\sum_{t=1}^{N}\left|y_t-\hat{y}_t\right|, \\
&\mathrm{MAPE}= \frac{1}{N}\sum_{t=1}^{N}
\left|\frac{y_t-\hat{y}_t}{y_t}\right| \times 100\%, 
\end{split}
\nonumber
\end{equation}
where $\{\hat{y}_i\}_{i=1}^{N}$ and $\{y_i\}_{i=1}^{N}$ denote the predictions and  the ground truth values, respectively. It is noted that for MAPE calculation a mask was applied on the raw ground truth values to filter out zero and near-zero values.

\paragraph{Reproducibility:} The source code, dataset, and evaluation scripts associated with this work are publicly available in our project repository\footnote{\url{https://github.com/yw3576/UniST-Pred}}. 

\subsection{Performance Comparison}

\paragraph{Quantitative Evaluation:}
Table~\ref{tab:comparison} reports the quantitative comparison results. On SimSF-Bay, UniST-Pred consistently outperforms all baselines, achieving the lowest errors across all evaluation metrics, while TS-Mixer achieves the second best RMSE.
On PEMS-Bay, UniST-Pred performs on par with the strongest baseline, STEP, in terms of RMSE, and remains close to the best MAE (1.91 versus 1.79 for STEP). This indicates that the proposed method achieves performance comparable to the leading baseline on this dataset, while its MAPE remains competitive, though lagging behind STEP.
On NYC-Taxi, UniST-Pred achieves the lowest RMSE and MAE, with noticeable margins over the closest baseline. Again, TS-Mixer achieves the second best RMSE, while STEP attains better MAPE, with UniST-Pred ranking second.
Taken together, these results indicate that UniST-Pred consistently captures absolute error patterns, as reflected by having best results in RMSE and MAE across datasets. 

\begin{figure}[!t]
\captionsetup{skip=5pt}
    \centering
    \includegraphics[width=1.0\columnwidth]{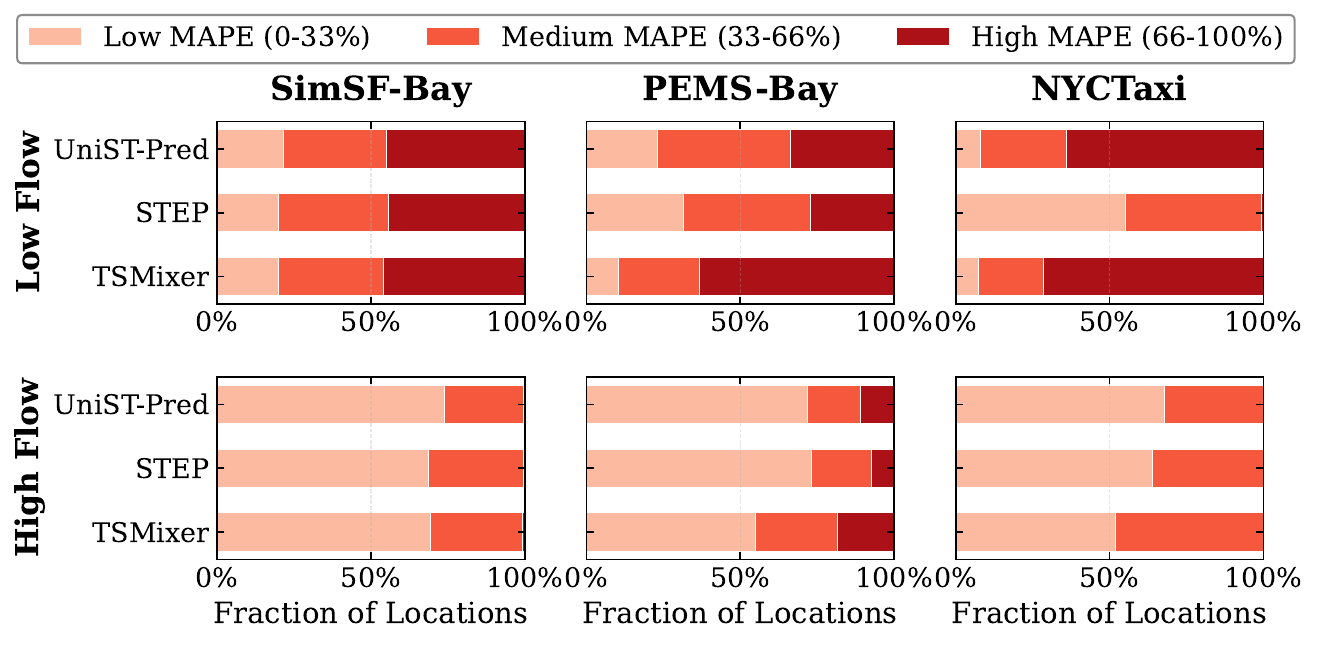} % Do not include the file extension
    \caption{Per-location MAPE distribution for UniST-Pred, STEP, and TSMixer on three datasets, stratified by high and low traffic flow. (Light colors indicate lower error.)}
    \label{fig:quanlitative}
\end{figure}

While average accuracy reflects overall performance, it does not show whether gains are consistent across locations or driven by a subset of easy cases. To examine spatial error patterns, Fig.~\ref{fig:quanlitative} reports per-location MAPE distributions for TSMixer, STEP, and UniST-Pred across all datasets. Locations are divided by traffic flow percentiles, with \textbf{high flow} locations in the top 25\% and \textbf{low flow} locations in the lower 75\%. Each stacked bar shows the proportion of locations with low (0--33\%), medium (33--66\%), or high (66--100\%) MAPE.

%\looseness-2

On SimSF-Bay, UniST-Pred achieves lower error on more locations than STEP and TSMixer, for both high-flow and low-flow locations. On PEMS-Bay and NYCTaxi, STEP shows a clear advantage on low-flow locations, while UniST-Pred matches or outperforms STEP on high-flow locations. These results suggest that UniST-Pred performs particularly well on high-traffic locations, which are often the most operationally important for traffic management applications. %\looseness-2

\begin{figure}[!t]
    \centering
\includegraphics[width=1.0\columnwidth]{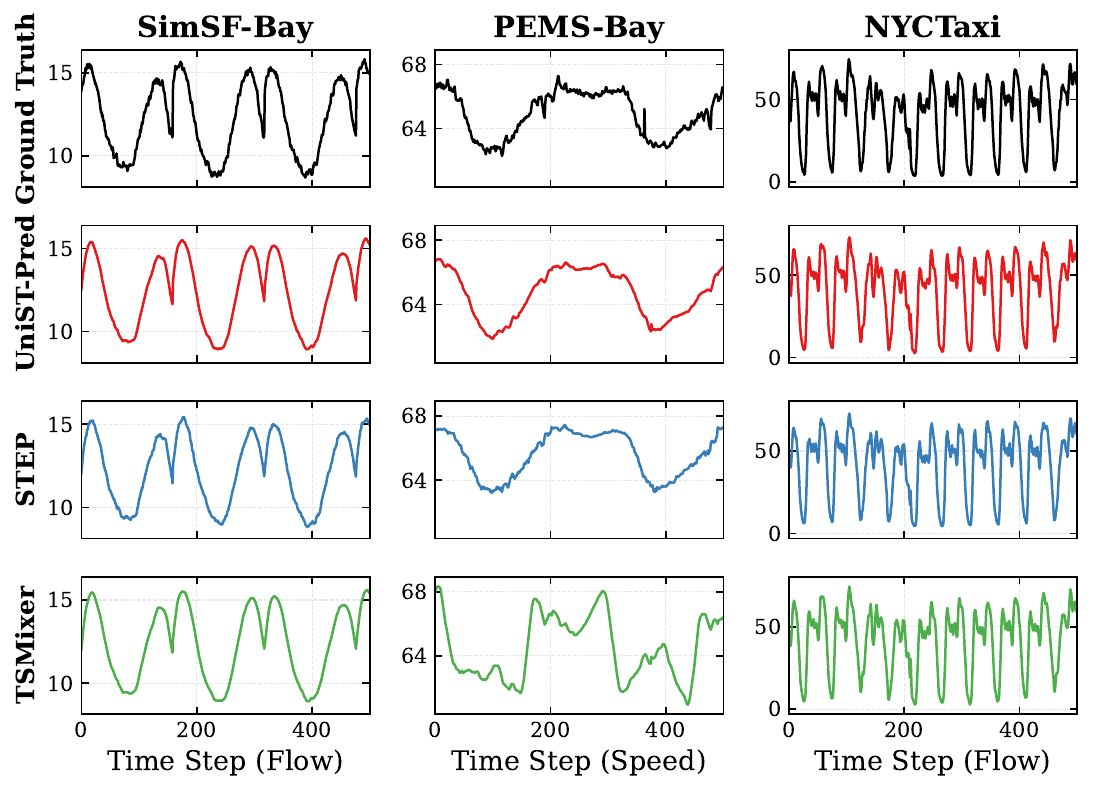} % Do not include the file extension
    \caption{Qualitative comparison of ground truth and model predictions.}
    \label{fig:comparison}
\end{figure}

\paragraph{Qualitative Evaluation:}
To examine prediction quality, Fig.~\ref{fig:comparison} shows the ground truth and predictions over all roads for three datasets, comparing UniST-Pred with STEP and TSMixer on a representative sample period. For SimSF-Bay, all methods capture the main periodic pattern. In PEMS-Bay, UniST-Pred tracks the broad pattern of the averaged speed and preserves the overall level and long-range trend, whereas STEP and TSMixer exhibit larger deviations near abrupt local changes. For NYCTaxi, all methods closely follow the ground truth across cycles, with UniST-Pred showing slightly higher fidelity, while the other methods exhibit slightly increased smoothing or variability. These results suggest that UniST-Pred (i) captures dominant temporal dynamics with better phase alignment, (ii) maintains stable predictions without noticeable drift over long horizons, and (iii) provides smooth and consistent estimates of the global signal, which is beneficial when the objective is accurate trend-level forecasting on network-aggregated traffic patterns.

\paragraph{Model Size:} Next, we compare the number of model parameters of UniST-Pred against STEP, one of the strongest baseline methods. As shown in Table~\ref{tab:model_parameters}, UniST-Pred is substantially more parameter efficient than STEP across all three datasets. This reduction is particularly noticeable on SimSF-Bay and PEMS-Bay, where our model is approximately 4.0$\times$ and 3.7$\times$ smaller than STEP, respectively, and becomes especially significant on NYC-Taxi, where UniST-Pred is over 30$\times$ smaller. This consistent reduction indicates that the proposed architecture achieves a favorable accuracy to efficiency trade-off, making it more suitable for deployment under computational constraints while maintaining strong predictive performance.

\begin{table}[!htb]
\centering
\resizebox{\columnwidth}{!}{
\begin{tabular}{lrrr}
\hline
\textbf{Method} & \textbf{SimSF-Bay} & \textbf{PEMS-Bay} & \textbf{NYCTaxi} \\
\hline
STEP  & 12,952,470 & 61,138,261 & 5,670,166 \\
UniST &  3,208,208 & 16,558,467 &   167,994 \\
Reduction (\%) & 75.23 & 72.91 & 97.04 \\
\hline
\end{tabular}
}
\caption{Comparison of model parameter counts between STEP and UniST across datasets.}
\label{tab:model_parameters}
\end{table}

\paragraph{Multi-horizon Forecasting:} We additionally evaluate UniST-Pred on SimSF-Bay under varying prediction horizons, specifically $H'=3,6,9$. The results, listed in Table~\ref{tab:multi-horizon}, show that UniST-Pred consistently achieves the lowest error metrics across all considered horizons compared with the benchmark methods. This suggests that the proposed approach generalizes well to multi-horizon forecasting settings while preserving strong predictive performance.

\begin{table}[!htb]
\centering
\resizebox{1.0\linewidth}{!}{
\begin{tabular}{ccccc}
\hline
 &  & \multirow{2}{*}{\textbf{TS-Mixer}} & \multirow{2}{*}{\textbf{STEP}} & \multirow{2}{*}{\textbf{UniST-Pred}}\\
\textbf{Horizon}  &  &  &  & \\
\hline
\multirow{3}{*}{$H'=3$} & RMSE & 4.14 & 4.38 & \textbf{3.73}  \\
 & MAE & 2.67 & 2.84 & \textbf{2.40} \\
 & MAPE (\%) & 38.23 & 41.28 & \textbf{33.29} \\
 \hline
\multirow{3}{*}{$H'=6$} & RMSE & 4.33 & 4.52 & \textbf{3.89}\\
 & MAE & 2.96 & 2.92 & \textbf{2.49} \\
 & MAPE (\%) & 40.47 & 41.92 & \textbf{34.27} \\
 \hline
\multirow{3}{*}{$H'=9$} & RMSE & 4.56 & 4.75 & \textbf{4.06} \\
 & MAE & 3.01 & 3.06 & \textbf{2.60} \\
 & MAPE (\%) & 43.01 & 43.62 & \textbf{35.37} \\
\hline
\end{tabular}
}
\caption{Performance comparison of TS-Mixer, STEP and UniST-Pred on SimSF-Bay across different forecasting horizons.}
\label{tab:multi-horizon}
\end{table}

\subsection{Ablation Studies}
\begin{table}[!b]
\centering
\resizebox{\linewidth}{!}{
\begin{tabular}{lccc}
\toprule
Method & RMSE $\downarrow$ & MAE $\downarrow$ & MAPE (\%) $\downarrow$ \\
\midrule
UniST-Pred (full) & \textbf{3.61} & \textbf{2.37} & \textbf{34.2} \\
\midrule
\multicolumn{4}{l}{\textit{Component removal}} \\
\quad w/o spatial     & 6.06 & 3.43 & 65.5 \\
\quad w/o temporal    & 5.64 & 4.11 & 49.8 \\
\quad w/o fusion      & 4.06 & 2.57 & 36.2 \\
\midrule
\multicolumn{4}{l}{\textit{Component replacement}} \\
\quad spatial $\rightarrow$ GCN   & 4.17 & 2.65 & 71.2 \\
\quad temporal $\rightarrow$ FC  & 3.96 & 2.58 & 38.1 \\
\bottomrule
\end{tabular}
}
\caption{Ablation results of UniST-Pred on SimSF-Bay.}
\label{tab:ablation}
\end{table}

To analyze the contribution of individual components in the proposed architecture, we conduct an ablation study by systematically removing or replacing key sub-modules. Specifically, we consider variants that remove the spatial block, the temporal block, or the fusion block, as well as variants that replace the spatial block with standard GCN layers or the temporal block with a fully connected (FC) layers.

Table~\ref{tab:ablation} presents the ablation study results. We observe that removing either the \textit{spatial block} or the \textit{temporal block} consistently degrades performance across all metrics. Similarly, replacing the spatial block with standard GCN layers or the temporal block with fully connected layers increases the error. The \textit{w/o fusion} variant performs competitively among simplified settings, indicating that the backbone alone can learn informative representations. The full model achieves the lowest error across all metrics, confirming that all proposed components contribute to the best overall performance.

\subsection{Case Study: Robustness \& Explainability}

 \begin{figure*}[!ht]
    \centering
    \includegraphics[width=\linewidth]{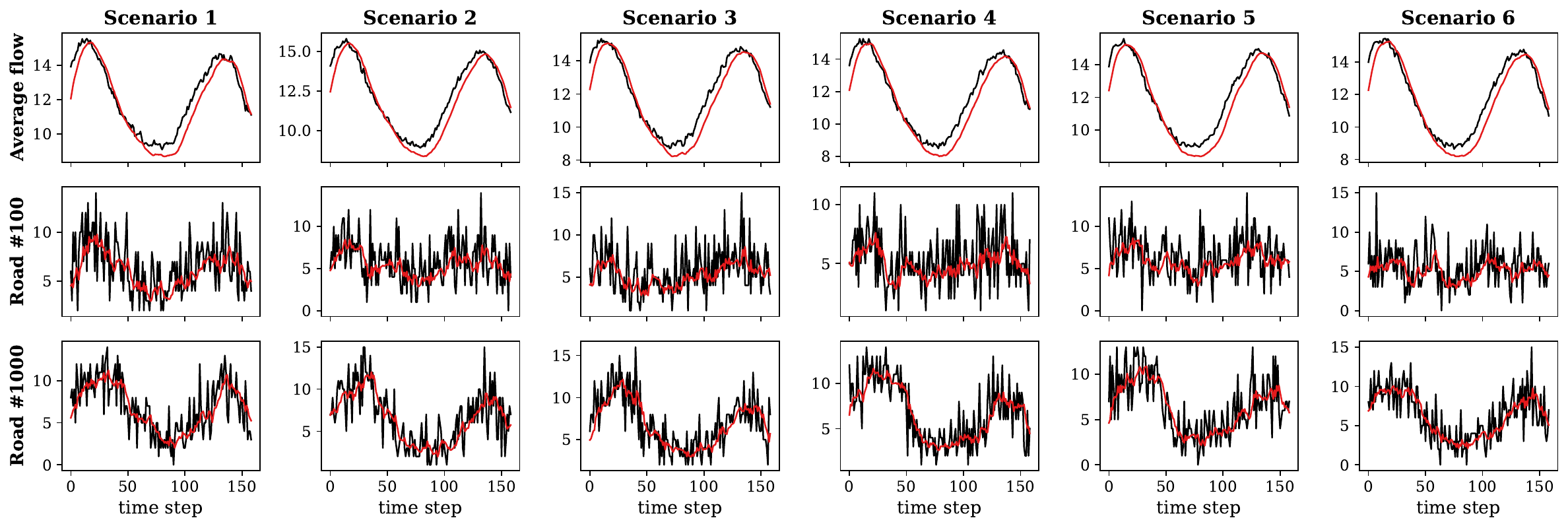} % Do not include the file extension
    \caption{Comparison of ground-truth traffic flow and UniST-Pred predictions across the six disruption scenarios, shown for the network-wide average (top row) and two representative roads (middle and bottom rows).}
    \label{fig:case study}
\end{figure*}

% Side-by-side figures
\begin{figure*}[!t]
%\captionsetup{skip=7pt}
\centering

\begin{subfigure}{0.31\textwidth}
  \centering
  \includegraphics[width=0.99\linewidth]{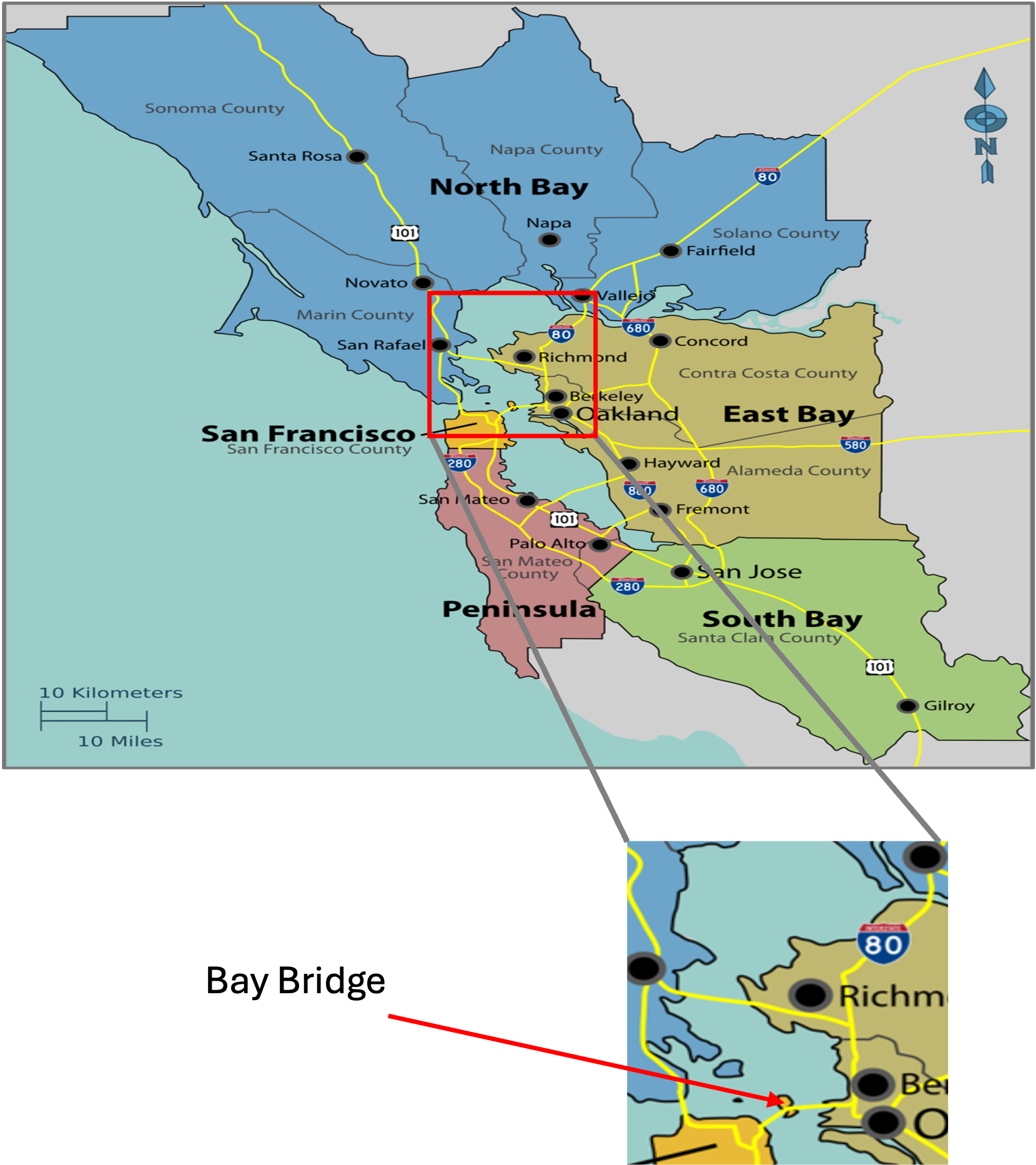}
  \caption{San Francisco Bay area}
\end{subfigure}%
\
\begin{subfigure}{0.31\textwidth}
  \centering
  \includegraphics[width=0.98\linewidth]{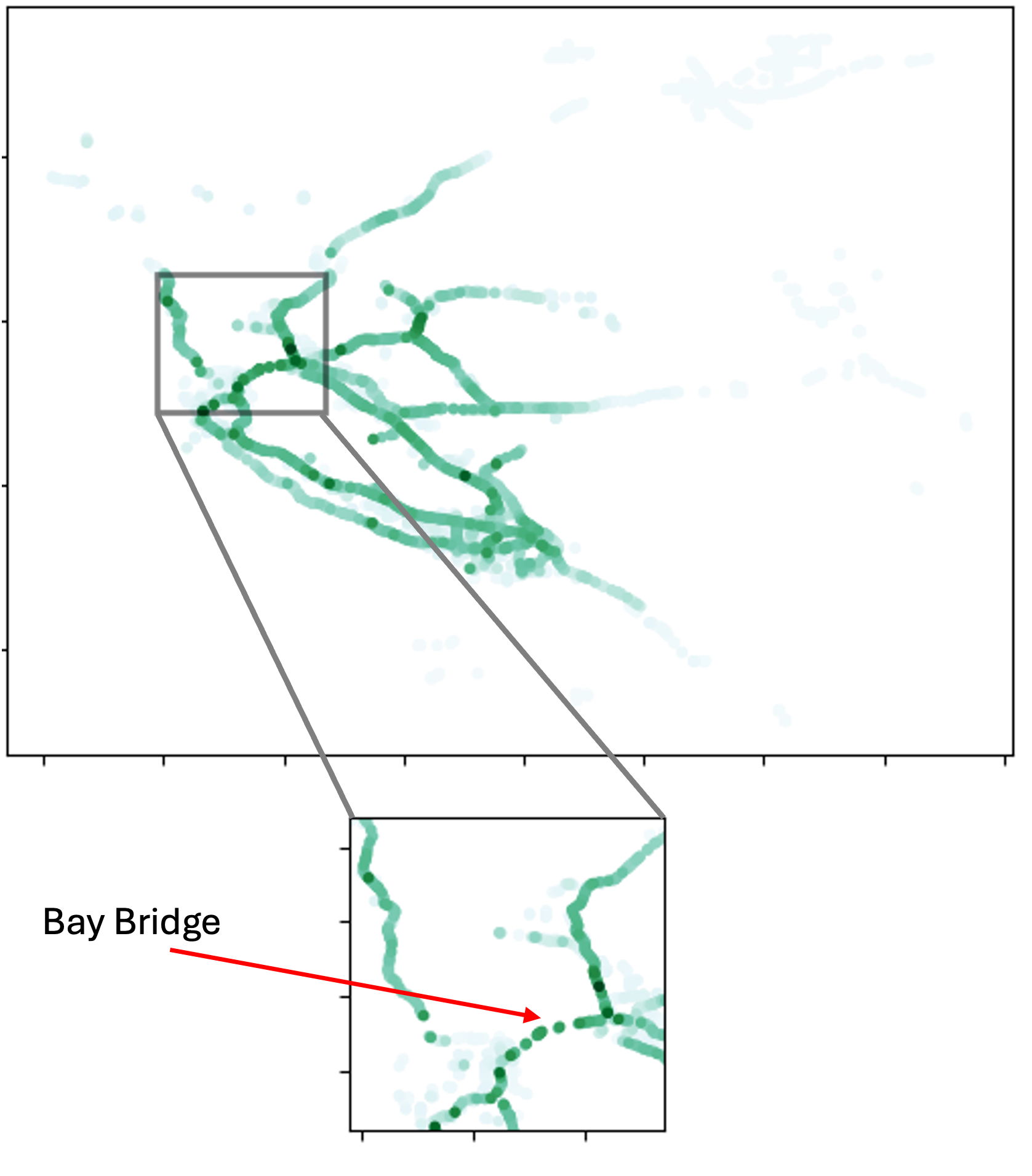}
  \caption{Fully connected network}
\end{subfigure}%
\
\begin{subfigure}{0.31\textwidth}
  \centering
  \includegraphics[width=.98\linewidth]{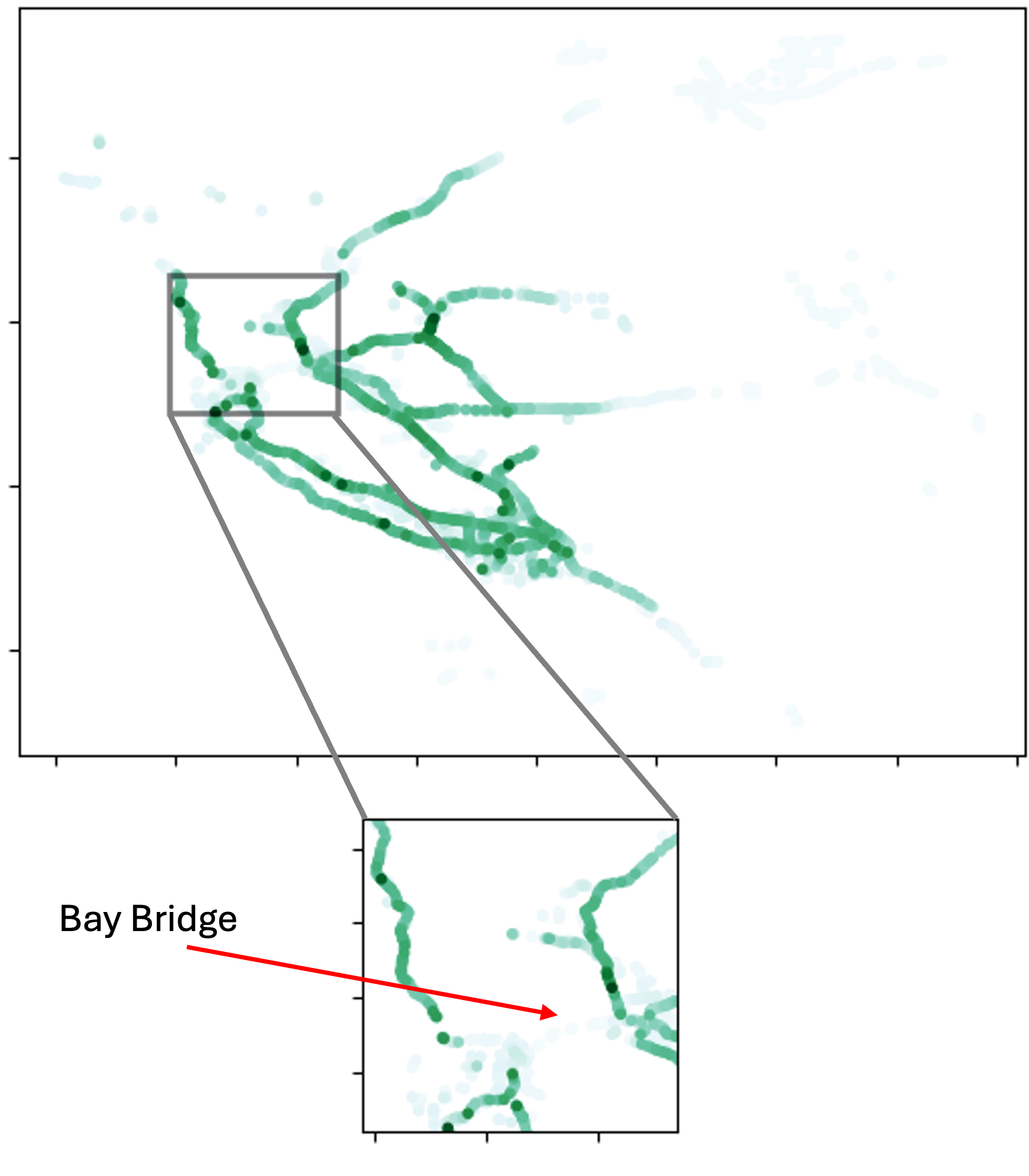}
  \caption{Scenario 1}
\end{subfigure}
\\
  
\begin{subfigure}{0.32\textwidth}
  \centering
  \includegraphics[width=.97\linewidth]{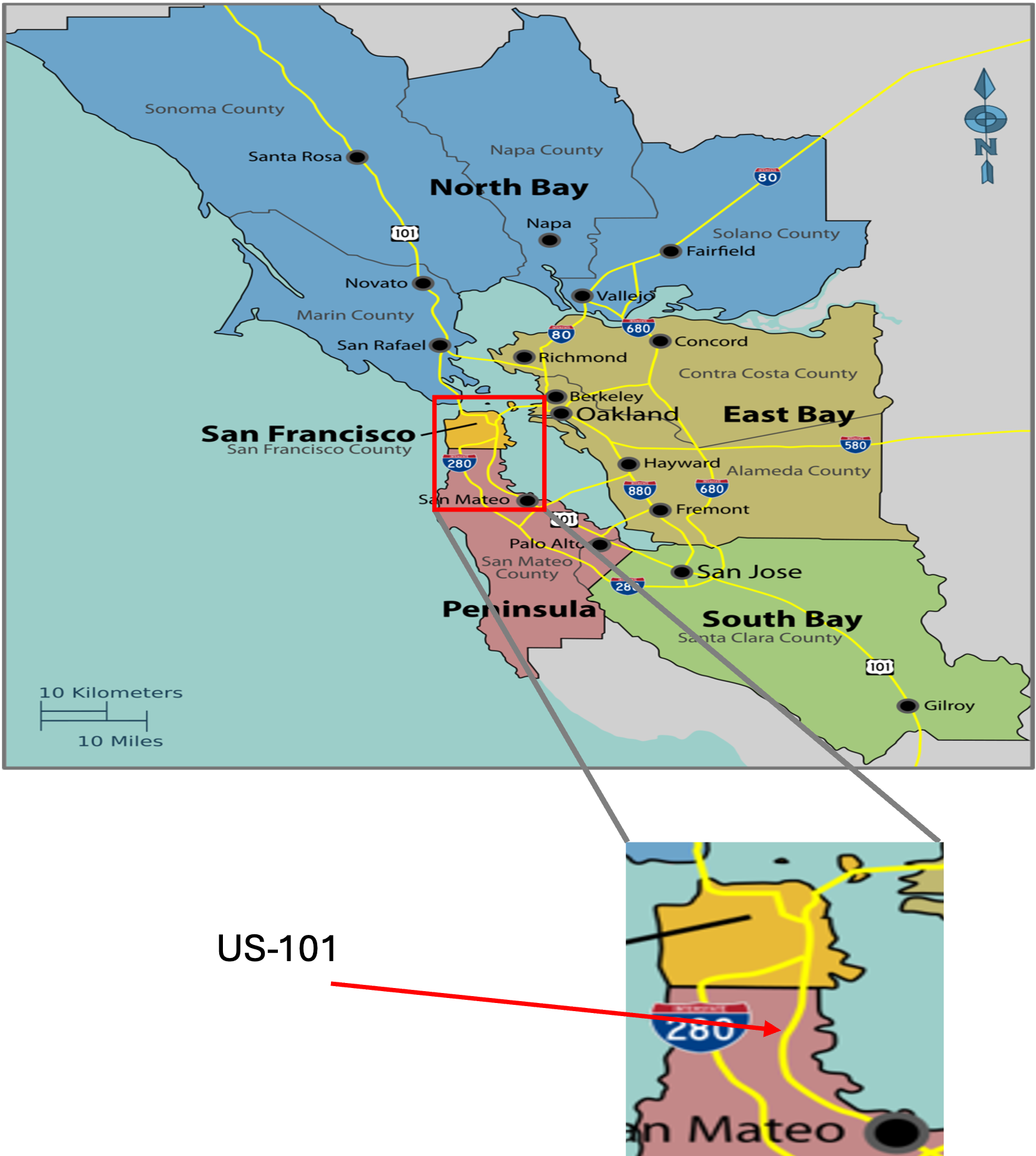}
  \caption{San Francisco Bay area}
\end{subfigure}
\
\begin{subfigure}{0.3\textwidth}
  \centering
  \includegraphics[width=.98\linewidth]{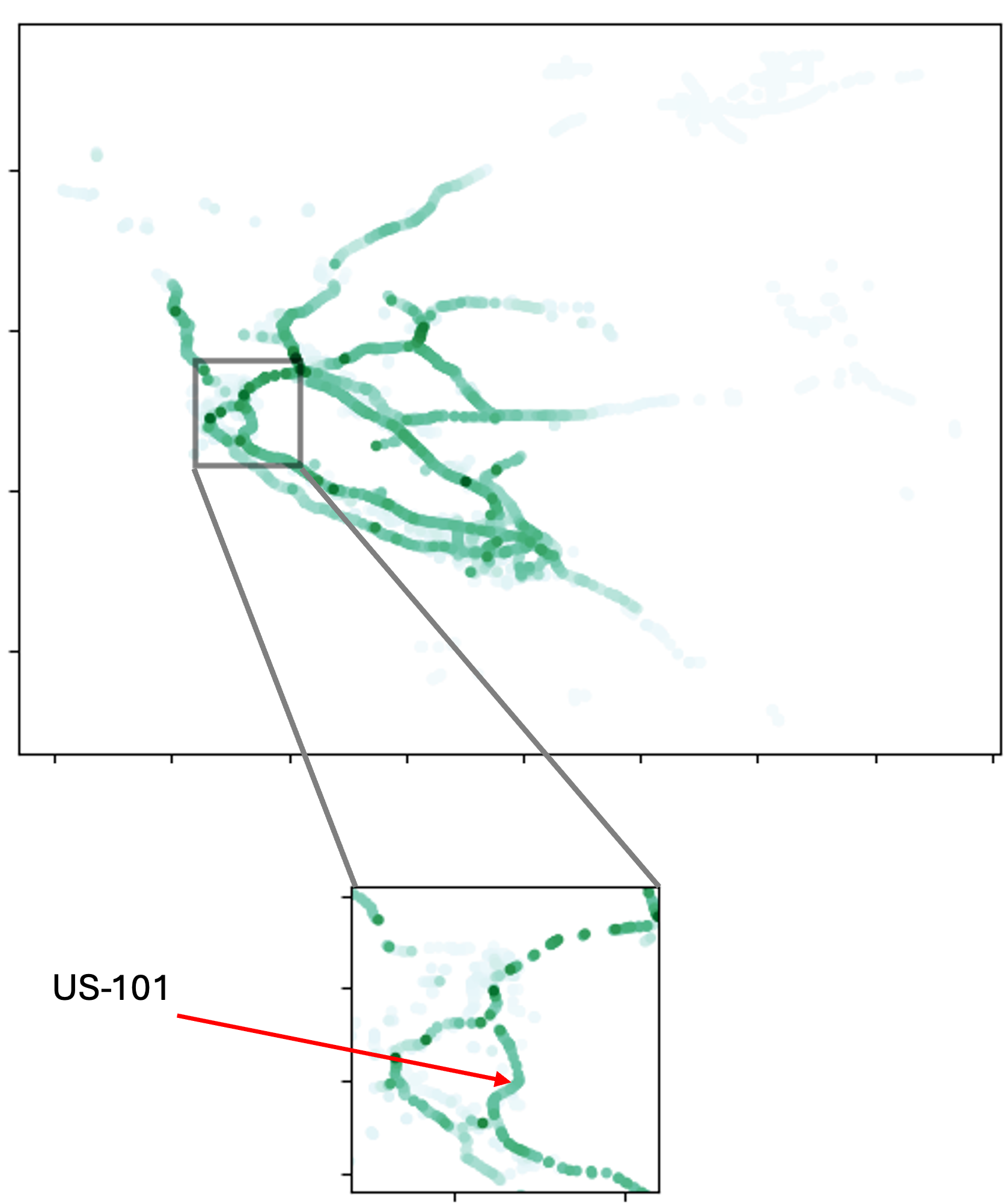}
  \caption{Fully connected network}
\end{subfigure}
\
\begin{subfigure}{0.3\textwidth}
  \centering
  \includegraphics[width=.98\linewidth]{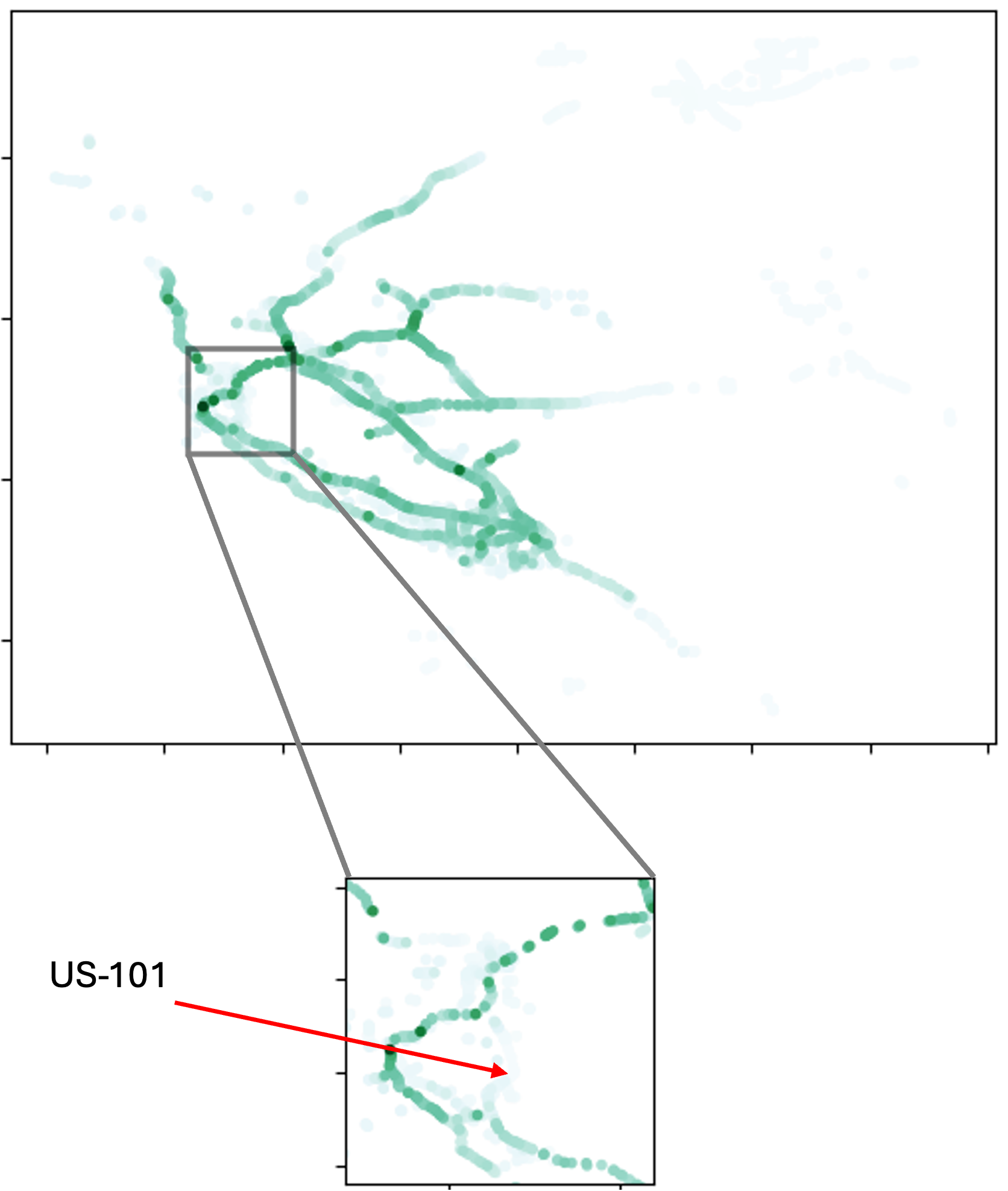}
  \caption{Scenario 4}
\end{subfigure}
\
\caption{{Explainability map for (c) Scenario 1: disconnecting Oakland Bay Bridge, (f) Scenario 4: disconnecting US-101, (b) and (e): all connected network. The zoom-in parts in (a) (b) (c) are the Oakland Bay Bridge and its surrounding area. The zoom-in parts in (d) (e) (f) are US-101 and its surrounding area. Darker color indicates higher contribution to the prediction.}}
\label{fig:explainability}
\end{figure*}

We conclude the empirical evaluation by examining the robustness of UniST-Pred under major infrastructure disruptions. To this end, we use the adopted MATSim model to construct six stress-test scenarios on the SF Bay Area network by removing major bridges or freeway/highway links that play a critical role in the region's connectivity. Specifically, we consider the removal/closing of \textbf{(Scenario 1)} Oakland Bay Bridge; \textbf{(Scenario 2)}  Carquinez Bridge and San Rafael Bridge; \textbf{(Scenario 3)}  Golden Gate Bridge and Dumbarton Bridge; \textbf{(Scenario 4)} US-101 highway (near Brisbane); \textbf{(Scenario 5)} the shared segment of I-80 and I-580 freeways near Albany; and \textbf{(Scenario 6)} the links considered in Scenarios 4 and 5. \looseness-2

\paragraph{Robustness Under Disruption:}
Fig.~\ref{fig:case study} compares the ground truth traffic flow and the predictions produced by UniST-Pred, both at the level of the network-wide average and for two representative roads, across all six scenarios. At the aggregate level (top row), the predictions closely track the ground truth. For individual roads (middle and bottom rows), the predictions remain trend-consistent, though brief local spikes may be smoothed or missed.

A quantitative comparison against TSMixer and STEP is reported in Table~\ref{tab:scenarios}, which shows that UniST-Pred attains the lowest RMSE in all six disruption scenarios as well as in the fully connected setting, indicating consistently strong performance across different connectivity conditions.

\begin{table}[bht]
\centering
\resizebox{\linewidth}{!}{
\begin{tabular}{cccc}
\hline
 & \textbf{TS-Mixer} & \textbf{STEP} & \textbf{UniST-Pred} \\
\hline
Scenario 1 & 3.89 & 3.93 & \textbf{3.61} \\
Scenario 2 & 3.87 & 3.95 & \textbf{3.59} \\
Scenario 3 & 3.87 & 3.91 & \textbf{3.60} \\
Scenario 4 & 3.89 & 3.91 & \textbf{3.59} \\
Scenario 5 & 3.87 & 3.91 & \textbf{3.58} \\
Scenario 6 & 3.88 & 3.90 & \textbf{3.58} \\
Fully connected & 3.90 & 3.89 & \textbf{3.62} \\
\hline
\end{tabular}
}
\caption{RMSE of UniST-Pred, TSMixer, and STEP across the disruption scenarios and the fully connected network.}
\label{tab:scenarios}
\end{table}

\paragraph{Explainability Analysis:}
Complementary to the above, we study the attribution behavior of UniST-Pred under disruption scenarios by examining which road segments contribute most to its predictions. To this end, we first 
leverage the trained UniST-Pred model to
generate traffic predictions, then apply Integrated Gradients (IG), which quantifies the contribution of each input feature by integrating gradients along a straight-line path from a baseline $\mathbf{x}'$ to the input $\mathbf{x}$. More concretely, for a differentiable scalar output $F(\mathbf{x})$, the IG attribution for feature $i$ is calculated as
$
\mathrm{IG}_i(\mathbf{x}) = (x_i - x'_i)\int_{0}^{1}\frac{\partial F\big(\mathbf{x}' + \alpha(\mathbf{x}-\mathbf{x}')\big)}{\partial x_i}\,d\alpha.
$

Fig.~\ref{fig:explainability} visualizes the resulting road-level IG contribution maps, where darker segments correspond to higher contribution to the model’s prediction. The figure contrasts the fully connected network (middle panel) with the disrupted cases in Scenario 1 and Scenario 4 (right panels), revealing a clear redistribution of attribution under disruption. Specifically, in the disruption scenarios, the model’s attribution shifts away from disconnected links toward alternative corridors, suggesting that the prediction mechanism adapts to the altered network topology. \looseness-2

\section{Conclusion}
\label{sec:conclusion}

This work studies spatio-temporal traffic forecasting as a modular modeling problem and proposes a unified framework that decouples temporal dependency learning from spatial representation extraction with adaptive fusion. By avoiding tightly coupled spatio-temporal propagation, the framework provides a lightweight yet expressive alternative for traffic prediction. Experiments on multiple benchmarks show that the proposed design achieves competitive performance with a compact architecture, while additional analyses offer insights into the contributions of temporal and spatial components. More broadly, this study suggests that a \emph{decouple-then-fuse} strategy can serve as a useful design principle for spatio-temporal learning, with relevance beyond traffic forecasting and practical importance for real world traffic networks.

%\section*{Ethical Statement}
%There are no ethical issues.

%\nocite{*}
%\section*{Acknowledgments}
%\textcolor{red}{This work was jointly supported by }

%supplementary material for IJCAI
%\clearpage
%\onecolumn
\appendix
\section{MATSim Simulator \& SF Bay Transportation Network Model}
\label{sec:matsim_dataset}

MATSim (Multi-Agent Transport Simulation)\cite{w2016multi} is a microscopic, agent-based transportation simulator widely utilized in transportation planning and mobility research. MATSim models travel demand explicitly at the individual level, representing each traveler as an autonomous agent with daily activity plans that evolve through iterative re-planning and network loading. At each iteration, agents execute their plans on a detailed transportation network, generating time-resolved link-level traffic states such as flow, speed, and travel time, which are subsequently used to update agent plans based on experienced costs and utilities defined by user-assigned scoring functions. The simulation is considered converged once agents' plans and experienced costs stabilize across successive iterations, corresponding to a stochastic user equilibrium. Owing to its realism  and flexibility, MATSim is commonly employed by academic researchers, transportation authorities, and planning agencies for scenario analysis, policy evaluation, and infrastructure planning.

Within the scope of current study, we adopt from  \cite{PAPAKONSTANTINOU2020102515} the MATSim model calibrated for the San Francisco Bay Area. In this model, the commuters travel plan is based on the vehicular travel information from Pozdnoukhov et al. \cite{pozdnoukhov2016san}. The demand model is based on anonymised cellular network infrastructure data stream and census data from the 2010-2012 California Household Travel Survey data. The demand generation relies on the 1454 Traffic Analysis Zones in the area developed by the Metropolitan Transportation Commission (see \url{https://abag.ca.gov/sites/default/files/pba_2050-regional_growth_forecast_methodology.pdf}). The model considers  a sample population of 463,938 commuters and the road links in the transportation network are scaled down to 8\% of their original capacities to correctly match the population scale.

\section{Details of Datasets}
\label{sec:matsim_dataset_2}

\begin{table}[!htbp]
\centering
\resizebox{\columnwidth}{!}{%
\begin{tabular}{lccc}
\toprule
\textbf{Property} & \textbf{PEMS-BAY} & \textbf{NYC Taxi} & \textbf{SimSF-Bay} \\
\midrule
Traffic variable          & Speed             & Flow    & Flow  \\
Spatial structure  & Sensor graph      & Fixed zones / grid & Dynamic road network \\
\# of nodes            & 325               & 200                & 7709 \\
\# of edges            & 2369              & 712                & 8781 \\
\# of time steps       & 5.2k+           & 2.6m+                & 168 (per scenario) \\
Time interval       & 5 min             & 30 min                & 5 min \\
Time range          & 01/01/2017 - 05/31/2017 & 01/01/2015 to 03/01/2015           & 6 AM to 8 PM (per scenario) \\
Network topology  & Fixed             & Fixed             & Varies across scenarios \\
\bottomrule
\end{tabular}
}
\caption{Statistics of datasets used in this study.}
\label{tab:dataset_stats}
\end{table}

\section{Model Parameters \& Implementation Details}
For the proposed UniST-Pred, the model and implementation details are listed in Table \ref{tab:params_impl_combined}. ARIMA was implemented via statsmodels python package \cite{Skipper_statsmodels_Econometric_and_2010}. LSTM was implemented from Pytorch Neural Networks package \cite{pytorch_website}. Other benchmarks were implemented via the code provided in their project repositories: TCN \cite{lea2017temporal}, TSMixer and its extended variant TSMixer-ext \cite{chen2023tsmixer}, STGCN \cite{yu2018stgcn}, ASTGCN \cite{guo2019attention}, ST-SSL \cite{ji2023spatio}, DCRNN \cite{li2017diffusion}, and STEP \cite{shao2022pre}. For LSTM and TCN, the hidden dimensions were selected via grid search. For TSMixer, we adopted the hyperparameter search procedure provided in its repository. For the graph-based baselines, we used the configurations released with their codebase. \looseness-2
\begin{table}[!htbp]
\centering
\resizebox{\columnwidth}{!}{
\begin{tabular}{lccc}
\hline
\textbf{Item} & \textbf{SimSF-Bay} & \textbf{PEMS-Bay} & \textbf{NYCTaxi} \\
\hline
\multicolumn{4}{l}{\textbf{Model Parameters}} \\
\hline
Input length ($H$) & 9 & 2016 & 35 \\
Prediction horizon ($H'$) & 1 & 12 & 1 \\
\hline
Number of channels ($C$) & 4 & 4 & 4 \\
Number of GT layers ($L_g$) & 2 & 2 & 2 \\
GCN in channel & 2 & 2 & 1 \\
GCN out channel & 100 & 100 & 100 \\
\hline
Number of TSMixer layers ($K$) & 2 & 4 & 2 \\
feature mixing MLP width & 100 & 100 & 100 \\
Dropout & 0.2 & 0.2 & 0.2 \\
\hline
SE reduction ratio ($r$) & 2 & 4 & 2 \\
\hline
\multicolumn{4}{l}{\textbf{Implementation Details}} \\
\hline
Hardware & \multicolumn{3}{c}{NVIDIA A100 80GB PCIe}\\
\hline
Optimizer & \multicolumn{3}{c}{ADAM} \\
Initial learning rate & 0.0005 & 0.0005 & 0.0005 \\
Weight decay & 0.1 & 0.1 & 0.1 \\
Batch size & 5 & 64 & 5 \\
Epochs & 20 & 50 & 100 \\

\hline
Loss function & MAE & Smoothed L1 & MAE \\
Train/val/test split & 66\%/17\%/17\% & 70\%/10\%/20\% & 70\%/10\%/20\% \\
\hline
\end{tabular}
}
\caption{Model parameters and implementation details}
\label{tab:params_impl_combined}
\end{table}

\clearpage

\section*{Acknowledgments}
The authors thank Madeline Sheehan for providing us with the data of the San Francisco Bay Area highway network and the code for processing it. This research was carried out on the High Performance Computing resources at New York University Abu Dhabi.

\bibliographystyle{named}
\bibliography{ijcai25}

\end{document}